\documentclass[conference]{IEEEtran}
\IEEEoverridecommandlockouts
\usepackage{cite}
\usepackage{hyperref}
\usepackage{amsmath,amssymb,amsfonts}
\usepackage{algorithmic}
\usepackage{graphicx}
\usepackage{enumitem}
\usepackage{textcomp}
\usepackage{xcolor}
\def\BibTeX{{\rm B\kern-.05em{\sc i\kern-.025em b}\kern-.08em
    T\kern-.1667em\lower.7ex\hbox{E}\kern-.125emX}}

\usepackage[absolute]{textpos}

\begin{document}
\begin{textblock}{5}(11.8,0.8)
(Special Session)
\end{textblock}
\IEEEoverridecommandlockouts
\IEEEpubid{\makebox[\columnwidth]{ 979-8-3503-6378-4/24\$31.00 \copyright2024 IEEE \hfill} \hspace{\columnsep}\makebox[\columnwidth]{ }}
\title{Reliable edge machine learning hardware for scientific applications\\
\thanks{This work is supported by the U.S. Department of Energy (DOE), Office of Science, Office of Advanced Scientific Computing Research under the ``Real-time Data Reduction Codesign at the Extreme Edge for Science'' Project (DE-FOA-0002501).
This material is based upon work supported by the National Science Foundation Graduate Research Fellowship Program under Grant No. DGE-2038238. 
Any opinions, findings, and conclusions or recommendations expressed in this material are those of the author(s) and do not necessarily reflect the views of the National Science Foundation.
}
}

\author{\IEEEauthorblockN{
Tommaso Baldi$^{\ast\dagger}$\\
Javier Campos$^\ast$\\
Ben Hawks$^\ast$\\
Jennifer Ngadiuba$^\ast$\\
Nhan Tran$^\ast$
}
\textit{$^\ast$Fermilab}\\
\textit{$^\dagger$University of Pisa}

\and
\IEEEauthorblockN{
Daniel Diaz\\
Javier Duarte\\
Ryan Kastner\\
Andres Meza\\
Melissa Quinnan\\
Olivia Weng
}
\textit{UC San Diego}
\and
\IEEEauthorblockN{
Caleb Geniesse \\
Amir Gholami \\
Michael W. Mahoney}
\textit{UC Berkeley/LBNL/ICSI}
\and
\IEEEauthorblockN{
Vladimir Loncar \\
Philip Harris}
\textit{MIT}
\and
\IEEEauthorblockN{
Joshua Agar \\
Shuyu Qin}
\textit{Drexel University}
}

\maketitle

\begin{abstract}
Extreme data rate scientific experiments create massive amounts of data that require efficient ML edge processing.  
This leads to unique validation challenges for VLSI implementations of ML algorithms: enabling bit-accurate functional simulations for performance validation in experimental software frameworks,  verifying those ML models are robust under extreme quantization and pruning, and enabling ultra-fine-grained model inspection for efficient fault tolerance. 
We 
discuss approaches to developing and validating reliable algorithms at the scientific edge under such strict latency, resource, power, and area requirements in extreme experimental environments.  
We study metrics for developing robust algorithms, present preliminary results and mitigation strategies, and conclude with an outlook of these and future directions of research towards the longer-term goal of developing autonomous scientific experimentation methods for accelerated scientific discovery.  
\end{abstract}



\section{Motivation}
\label{sec:motivation}

Ground-breaking science requires instruments that push sensing technology with increasing spatial and temporal resolution to explore nature at unprecedented scales and in extreme environments.  
This has led to a data generation explosion, with more and more data being generated in next-generation experiments.  
For example, particle physics experiments look for extremely rare collision events (one in a billion billion) that can answer fundamental questions about the fabric of spacetime or the nature of dark matter.  
Alternatively, microscopy experiments take hundreds of thousands of images per second to understand material properties that can advance computing, quantum science, and basic energy research. 
There are many other applications in a wide range of domain sciences, including fusion, nuclear physics, neuroscience, and quantum computing, that can benefit from real-time, low-latency edge processing~\cite{deiana2022applications}. 

\begin{figure}[t]
\centering
    \includegraphics[width=0.95\columnwidth]{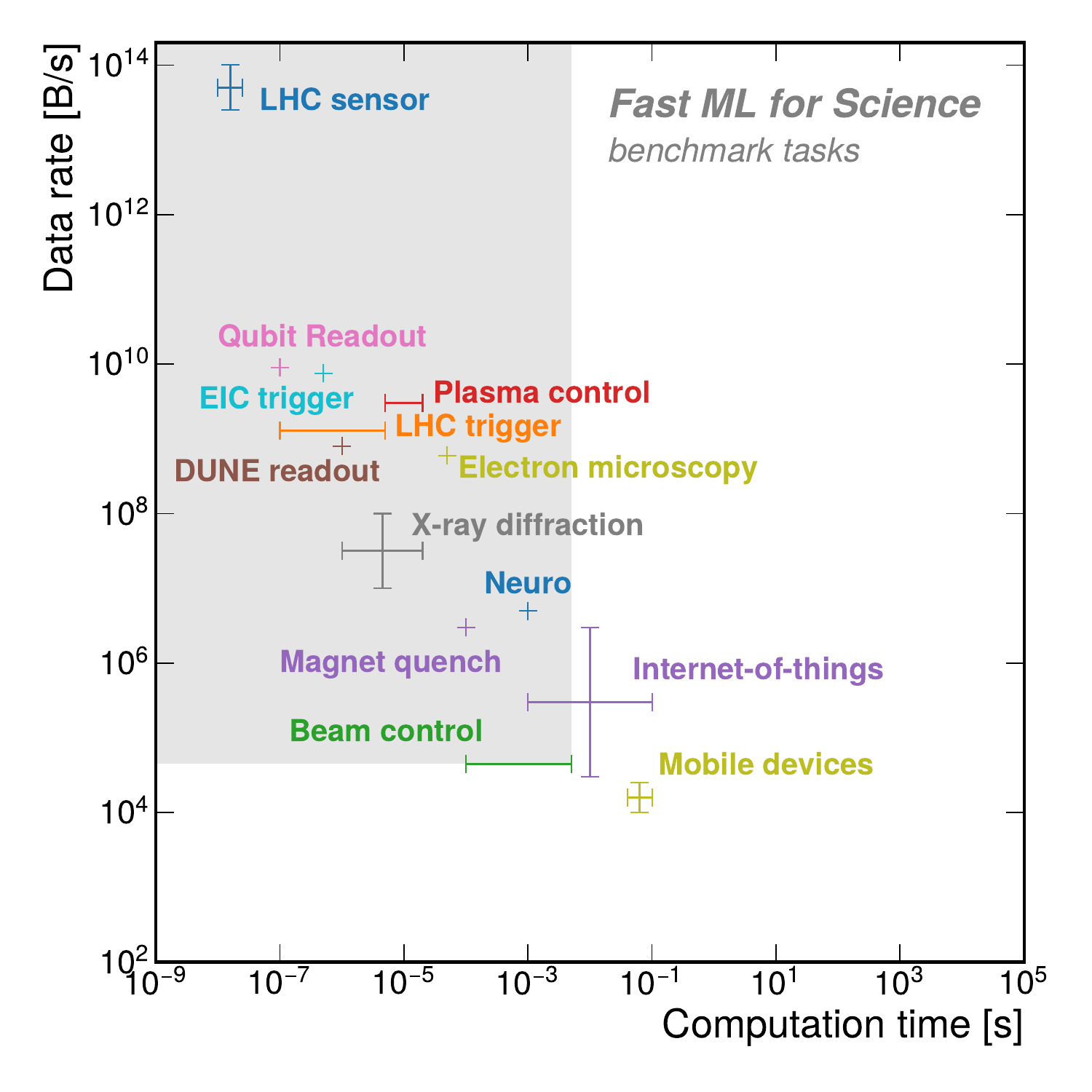}
    \caption{Many scientific and edge ML benchmark tasks~\cite{deiana2022applications} must process incoming data at a high rate leading to extreme low-latency and high-bandwidth requirements.  Applications illustrated here range across particle physics (LHC, DUNE), nuclear physics (EIC), material science (X-ray diffraction, microscopy), neuroscience, fusion energy, quantum information science, superconducting magnet research, and particle accelerators.  This can be compared against traditional internet-of-things and mobile device applications which are less stringent.}
\label{fig:sciml}
\vspace{-5mm}
\end{figure}

Many of these experiments create terabytes to petabytes of data per second, and at this rate these data cannot be stored and processed with traditional methods in off-the-shelf computing clusters.  
Instead, scientists must process the data as close as possible to the experimental sensor–--at the edge.
This is similar in some ways to autonomous vehicles and other smart sensing applications, but it occurs at unprecedented data rates and latency requirements for data processing.  
An illustration of this is provided in Fig.~\ref{fig:sciml}, which shows various benchmark tasks for different scientific applications as well as their input bandwidth and processing latency requirements.  

These challenges require efficient, specialized compute hardware at the extreme edge, e.g., FPGAs, ASICs, and systems on chip (SoCs).
These platforms are common to many scientific experiments.
This also has implications for the computer architectures~\cite{weng2024architectural} that need to meet these extreme low-latency requirements: (i) often all neural network (NN) parameters must fit on-chip; and (ii) fully on-chip inference often requires hardware-software codesign with custom/reconfigurable logic to meet latency and bandwidth constraints.
Per-sensor compression and efficient aggregation of information, while preserving scientific fidelity, can significantly impact experiment data flow, analysis, control, and operation, as well as, ultimately, how quickly experiments can be performed and hypotheses explored.

In this paper, we discuss various unique and important challenges posed by deploying edge ML algorithms in realistic scientific environments with unfiltered and dynamic data streams and present proof-of-principle case studies and methodologies that address those challenges. 
We examine techniques:
\begin{itemize}[leftmargin=8pt]
    \item 
    to validate algorithm performance in large experimental software frameworks on simulated and real data with functionally verified VLSI implementations;
    \item 
    to characterize the stability of both local and global loss landscape structures in the training of highly customized and optimized algorithms;
    and
    \item 
    to improve the tolerance of edge ML algorithms to bit flip and sensor noise faults caused by experimental conditions with ultra-fine-grained bit level inspection and targeted regularization techniques.
\end{itemize}




\section{Exemplar application and previous work}
\label{sec:app}

For our example domain science task, we consider the CERN Large Hadron Collider (LHC) Compact Muon Solenoid (CMS) experiment~\cite{chatrchyan2008cms}.
This is an experiment that runs particle collision experiments that generate data rates of $\sim$40\,TB/s.
To reduce data rates, physicists deploy tens of thousands of ASICs and hundreds of FPGAs to make decisions on whether a specific collision is of particular interest (known as triggering) at the microsecond scale~\cite{CMS:2020cmk,CERN-LHCC-2020-004}.
One such ASIC is the endcap concentrator (ECON-T) ASICs~\cite{di2021reconfigurable}, which is planned for deployment in the upcoming LHC upgrade.  
Each ECON-T ASIC is running a NN encoder to compress experimental data from the high-granularity endcap calorimeter (HGCAL)~\cite{cms2017phase} into a smaller format for easy filtering in the trigger system.
The ECON-T encoder hardware must accept new input data at 40\,MHz and complete inference in 25\,ns within an area budget of 4\,mm$^2$~\cite{di2021reconfigurable}.
To meet these constraints, the ECON-T encoder model is a small two-layer NN with $\sim$2000 parameters quantized to have 6-bit fixed-point weights which operates completely on-chip.
To complicate matters further, the ECON-Ts operate in a high-radiation environment (due to their close proximity to particle collisions in the LHC). 
High radiation causes transient hardware errors, which can lead to incorrect application output (silent data corruptions) if the hardware is not designed robustly.
The ECON-Ts filter terabytes per second of data for high-energy physics studies, and faulty execution is unacceptable.
Only the NN weight parameters are vulnerable to faults because the activations are not stored in on-chip memory for longer than a cycle, as inference completes in a single cycle. 



The open-source quantization-aware platform QKeras~\cite{Coelho:2020zfu} is used for model training required for the algorithm development and also for generating stimuli for verification. 
The output of this stage is processed by the hls4ml compiler~\cite{duarte2018fast,fastml_hls4ml}, which translates the QKeras model description into a resultant {\texttt{C++}} description. 
This is then converted to an hardware RTL description in Verilog utilizing Mentor Catapult HLS. 
Several verification steps are undertaken at this stage to identify bugs and improve performance: design rule checks, C simulation and code coverage, and other traditional on-chip digital verification.


The HLS model description requires approximately 1,000 lines of code.
This stage is fast, $\mathcal{O}$(seconds), but it requires several hundred iterations to optimize the algorithm performance based on physics metrics. 
The HLS stage determines the level of parallelism in the design, choice of pipelining, resource reuse factor, and clock frequency. 
This directly impacts the total area, power consumption, and the latency of the design. 
The digital simulation and implementation stages, alternatively, is much more time intensive and can take several orders of magnitude longer~\cite{di2021reconfigurable}.


\section{Methodology and metrics}


For ML models developed for scientific edge application, taking the ECON-T as an example, we now discuss methods developed for large scale, bit accurate simulation; metrics for robust quantized training; and approaches for mitigating effects of sensing and electronics faults.

\subsection{Accurate and fast functional simulation}

Powerful ML algorithms are valuable for real-time data processing, but they can lead to complex designs in hardware platforms such as FPGAs and ASICs. 
At the same time, for high data rate experiments, these algorithms need to be validated against large amounts of data offline to understand their performance in identifying rare and interesting signals amongst large background processes.  
This can necessitate a CPU-based inference of the model over millions or billions of events within the simulation framework of large experimental collaborations.  

To that end, a firmware generation workflow like hls4ml has an extremely useful benefit that was not intended in its initial design.  
The hls4ml workflow emits a fuctionally \textit{bit accurate} C representation of the NN in hardware with the aid of algorithmic C\footnote{\url{https://hlslibs.org/}} and arbitrary precision\footnote{\url{https://docs.amd.com/r/en-US/ug1399-vitis-hls/Arbitrary-Precision-AP-Data-Types}} libraries.
The generated C code contains no external dependencies and compiles with any modern C/C++ compiler, making it easy to integrate into existing simulation frameworks. 
This can be used to test the functional performance of the hardware algorithm in high level programming languages like Python and C much more easily and rapidly rather than in lower level hardware description languages.  

For CMS trigger applications, it is necessary to emulate the ASIC and FPGA algorithms in software in order to evaluate their performance in simulation and compare data with simulation.
The firmware simulation generated with hls4ml has been added to the CMS software framework (CMSSW)~\cite{CMS:2006myw,cmssw} used to collect, produce, and analyze physics data.
Being fully self-contained, the generated C representation is compiled as a shared library and dynamically loaded, enabling multiple model versions to simultaneously co-exist. 
This enables physicists to trust the performance of algorithms developed for hardware despite being tested on very different hardware platforms.  

\subsection{Quantized NN loss landscapes}

\begin{figure}[tbh]
    \centering
    \includegraphics[width=0.48\columnwidth]{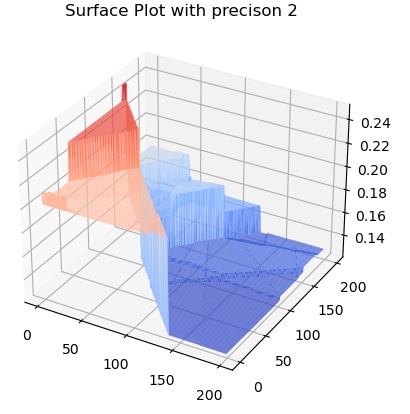}
    \includegraphics[width=0.48\columnwidth]{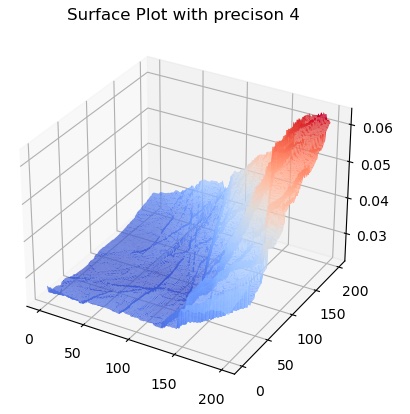}\\
    \includegraphics[width=0.48\columnwidth]{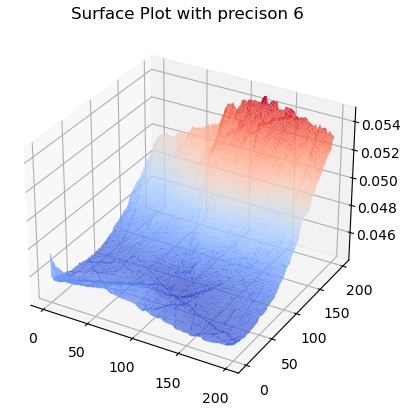}
    \includegraphics[width=0.48\columnwidth]{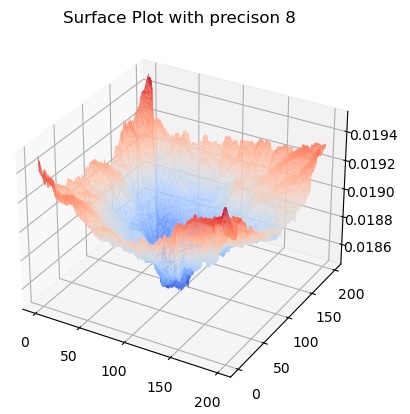}
    \caption{
        ECON-T model loss landscapes illustrating varying behaviors with different uniform quantizations between 2-bit and 8-bits.  A range of performance can be seen from very jagged landscapes at 2-bit weights to relatively smooth landscapes at 4- and 6-bit weights to a sharp narrow minima for 8-bit weights.   
    }
    \label{fig:loss}
\end{figure}

Bit accurate representations of these edge ML algorithms is important because of the custom quantization often deployed to make highly optimized implementations for efficient hardware.  
Often training is performed for such algorithms with a Pareto optimization over the performance and the system constraints (resources, latency, area, power, etc.) \cite{gholami2021survey}.  
However, the stability of the training is not typically considered, and the robustness of the model may depend on effects which characterize the local and global loss landscape structure. 
This can lead to unstable and unreliable model training and a high sensitivity to small perturbations to the input data  distribution. 
As an illustration of this, in Figure~\ref{fig:loss}, we show the loss landscapes of 4 ECON-T models trained with different uniform quantizations \cite{yao2020pyhessian, yang2021taxonomizing}. 

These visualizations help us to understand the local stability of each trained model under various quantizations.  To characterize these features, we explore a few metrics.
\begin{itemize}[leftmargin=8pt]
    \item \textbf{CKA similarity:} This metric, based on the centered kernel alignment (CKA), has shown effectiveness in capturing the correspondence between representations in various trained neural networks~\cite{kornblith2019similarity}, owing to its invariance properties. Models converging to the same local minima tend to demonstrate similar characteristics, offering insights into the smoothness of the loss landscape and the potential correlation with model performance and robustness. A low value of CKA suggests a high difference between models initialized with different random parameters, implying a challenging loss landscape.
    \item \textbf{Hessian}: In machine learning (ML), the Hessian matrix is a square matrix that characterizes the curvature of the loss function at a specific point. The eigenvalues of the Hessian matrix provide scalar values that offer insights into the curvature type at that point. Positive eigenvalues suggest local convexity of the function, indicating a single minimum or maximum. 
    \item \textbf{Neural Efficiency}: This metric, even if not directly related to the shape of the loss landscape, can give interesting insights about the capacity of the model. Indeed, network efficiency lets us know what is the percentage of node that are effectively used during the inference of the neural network, by estimating the minimum number of neurons required to encode the information exported by the neural layer, assuming perfect encoding~\cite{network-efficiency}.
\end{itemize}

\begin{figure}[tbh]
    \centering
    \includegraphics[width=0.49\columnwidth]{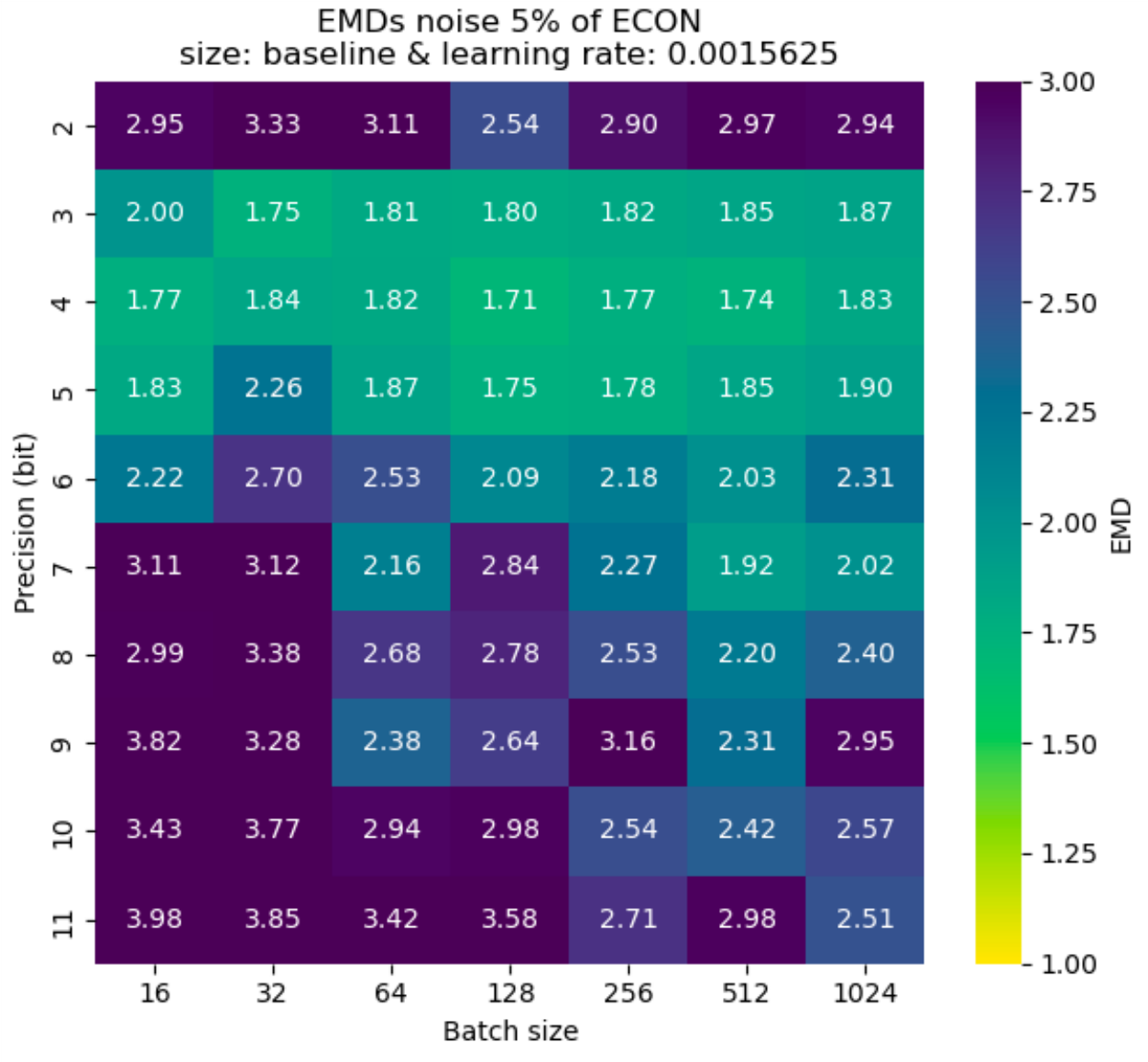}
    \includegraphics[width=0.47\columnwidth]{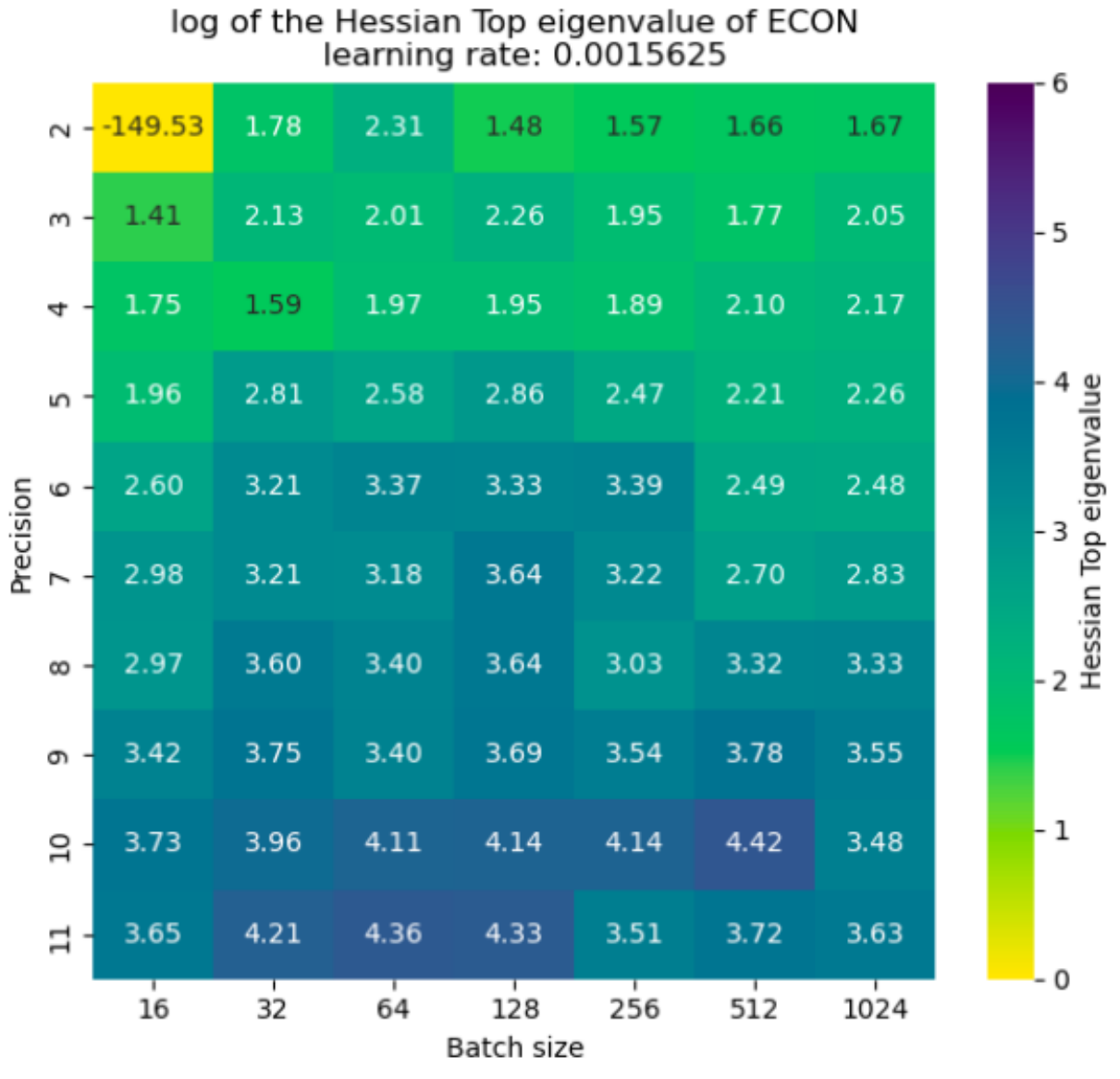}
    \caption{
    Results achieved by different versions, in terms of hyperparameters, of the ECON-T model. 
    The left heat map shows the EMD achieved by the model on noisy data. 
    The right heat map hows the top eigenvalue, in logarithmic scale. 
    }
    \label{fig:heatmaps}
\end{figure}

Figure~\ref{fig:heatmaps} shows the outcomes of the ECON-T analysis. 
Performance heat maps show the performance of the autoencoder by the Earth Mover's Distance metric when 5\% noise is introduced to the model inputs, Figure~\ref{fig:heatmaps}(left), revealing that the reduced precision of parameters in quantized NNs serves as regularization, guiding the model towards flat and smooth minima during training (as observed also in Figure~\ref{fig:loss}).  This is also reflected in the Hessian trace metric, Figure~\ref{fig:heatmaps}(right) which indicates lower bit widths have a flatter global minima.


\subsection{Fault tolerance to bit flips and sensor noise}

The final unique scientific challenge we consider is the sensitivity of the NN model to common faults: bit flip faults and sensor noise.

In Section~\ref{sec:app}, we discussed that the ECON-T ASIC will operate in an extreme radiation environment, 1000$\times$ that experienced in outer space.  
This leads to a large amount of single event upsets which cause bits to flip and which affect the performance of the ASIC.  
Typical radiation mitigation in hardware deploys triple modular redundancy (TMR) to triplicate the weight registers, but this incurs a 200\% resource overhead.  
In response, we develop FKeras~\cite{weng2023fkeras} a tool that performs ultra-fine-grained inspection of the model to study the bit-level sensitivity of each weight in the NN.  
This allows us to prioritize which bits need protection and which may be safely disregarded, reducing resource overhead.

We use FKeras~\cite{weng2023fkeras} to perform a design space exploration on the ECON-T model with respect to model size, performance, and fault tolerance.
We perform a neural architecture search of the ECON-T model, using a Bayesian optimization to find small, medium, and large quantized ECON-T models that range from less accurate to more accurate.
In our FKeras analysis, we find that all of the weight bits in the small, less accurate model are sensitive to faults, i.e., flipping any bit in the weights will cause the model to perform worse.
For the large, more accurate model however, we find that only 6\% of the weight bits are sensitive to faults.
This implies that we need only protect a small fraction of the weights~\cite{weng2023fkeras}. 
The tradeoff here is that we need more resources to implement a larger model.
As a result, we are met with a design tradeoff that merits careful consideration: would it be more worthwhile to implement a smaller, less accurate model that requires full protection or a larger, more accurate model that requires a small fraction of protection?

\begin{figure}[tbh]
    \centering
    \includegraphics[width=0.8\columnwidth]{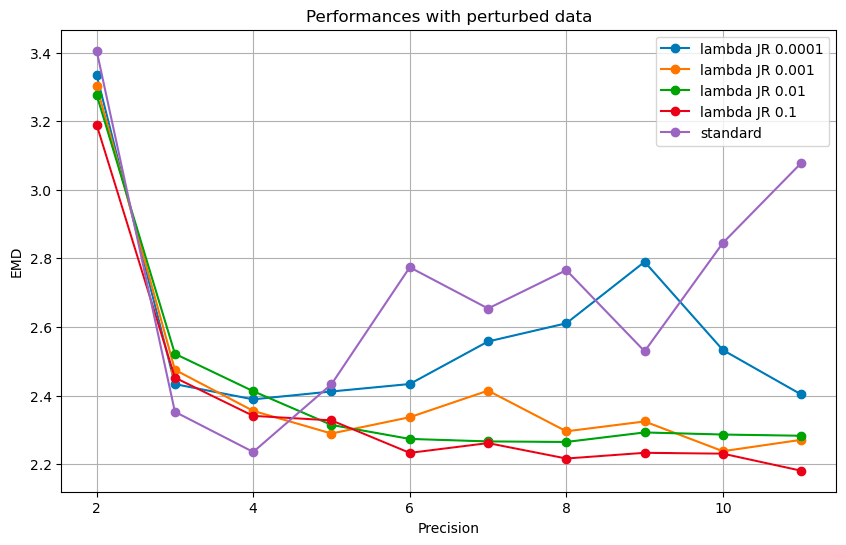}
    \caption{The plot illustrates the performance of ECON-T models under 5\% noisy data trained with different values of $\lambda_{JR}$, the hyperparameter used to tune the weight of the Jacobian regularization component}
    \label{fig:jreg}
\end{figure}

Another common fault model for sensing applications is noise which may come from the sensor itself or conditions in which it is taking data. Therefore, deploying models on the edge demands more than just achieving the desired performance metrics; it necessitates ensuring robustness to perturbations. Robust models can withstand and adapt to noisy environments, offering more reliable performance under varying input data. 
To meet this challenge, we employ specific mitigation techniques that leverage insights from the loss landscape analysis. The investigation into the ECON-T model highlighted the advantageous effect of converging towards wide and flat minima. To address this need, we introduced Jacobian Regularization into the training process, aiming to push decision boundaries further from individual data points by minimizing the Frobenius norm of the Jacobian matrix~\cite{jacobian}. As showed in Figure \ref{fig:jreg}, this regularization technique offers a targeted approach to enhancing model robustness, informed by the shape and characteristics of the loss landscape.

\section{Outlook}


In this paper, we highlight a number of unique challenges presented by high-throughput, low-latency scientific edge ML applications to enable reliable and efficient hardware VLSI implementations.  From accurate, large-scale simulation to characterizing quantization robustness to fault mitigation under bit flips and noise, we present methods that can be employed to ensure trustworthy and resilient operation of cutting-edge science experiments.  The methods we discuss in this study cut across a wide range of applications beyond particle physics.  Future studies will continue to study these methods on other scientific domains such as those presented in Fig.~\ref{fig:sciml}.  

The state-of-the-art of edge ML applications in science have made significant advances in performance and hardware optimization~\cite{deiana2022applications}. 
We discuss a number of important reliability challenges, though many remain, and we call attention to the importance of designing tools, metrics, and techniques to optimize for robustness as well which will enable (semi-)autonomous, adaptive, and intelligent experimentation.  Furthermore, visualization tools to aid domain scientists will enhance and transform the way experiments will operate in the future.  

From the studies presented here, several interesting research lines can be extended.  
We would like to explore how the fault tolerance of a neural network interacts with common neural network compression techniques like pruning and quantization. 
Perhaps certain pruning techniques lead to more resilient neural network weights, similar to a finding in which the chosen pruning technique made a neural network more resilient to noise in its inputs~\cite{diffenderfer2021winning}.
It also would be worthwhile to see how loss landscapes reveal information on how resilient a neural network is to faults.
While some correlation between robustness and loss landscape metrics were found, the extent of that correlation and possible causality still need to be understood.  Ultimately, the design space optimization of performance, robustness, and efficiency is still in early exploration for scientific ML applications.

\bibliographystyle{IEEEtran}
\bibliography{refs}

\end{document}